\documentclass[10pt,twocolumn,letterpaper]{article}
\usepackage[accsupp]{axessibility} 

\usepackage{cvpr}              
\usepackage{comment}

\usepackage[dvipsnames]{xcolor}

\definecolor{cvprblue}{rgb}{0.21,0.49,0.74}
\usepackage[pagebackref,breaklinks,colorlinks,citecolor=cvprblue]{hyperref}
\usepackage{amsmath}
\usepackage{pifont}
\usepackage{multirow}
\usepackage{arydshln}
\usepackage[normalem]{ulem}

\definecolor{junwen}{rgb}{0.9,0.5,0.1}
\definecolor{hao}{rgb}{0.6,0.3,0.9}
\newcommand{\hao}[1]{\textcolor{hao}{\emph{hao:~{#1}}}}

\definecolor{ben}{rgb}{0.9,0.,0.5}

\definecolor{slo}{rgb}{0.0,0.1,0.9}

\definecolor{ptr}{rgb}{0.3,0.5,0.1}

\newcommand{\OURS}{MatchU}

\def\paperID{5020} 
\def\confName{CVPR}
\def\confYear{2024}

\title{MatchU: Matching Unseen Objects for 6D Pose Estimation from RGB-D Images}

\author{Junwen Huang$^{1,2}$ \; Hao Yu$^1$ \;  Kuan-Ting Yu$^3$ \; Nassir Navab$^{1,2}$ \; Slobodan Ilic$^{1}$ \; Benjamin Busam$^{1,2,4}$  \\[0.5em]
$^{1}$ Technical University of Munich
\quad $^{2}$ Munich Center for Machine Learning\\
\qquad $^{3}$ XYZ Robotics
\qquad $^{4}$ 3dwe.ai
}

\begin{document}
\maketitle
\begin{abstract}
Recent learning methods for object pose estimation
require resource-intensive training for each individual object instance or category, hampering their scalability in real applications when confronted with previously unseen objects. 
In this paper, we propose \OURS{}, a Fuse-Describe-Match strategy for 6D pose estimation from RGB-D images. \OURS{} is a generic approach that fuses 2D texture and 3D geometric cues for 6D pose prediction of unseen objects. 
We rely on learning geometric 3D descriptors that are rotation-invariant by design. By encoding pose-agnostic geometry, the learned descriptors naturally generalize to unseen objects and capture symmetries. To tackle ambiguous associations using 3D geometry only, we fuse additional RGB information into our descriptor. This is achieved through a novel attention-based mechanism that fuses cross-modal information, together with a matching loss that leverages the latent space learned from RGB data to guide the descriptor learning process. 
Extensive experiments reveal the generalizability of both the RGB-D fusion strategy as well as the descriptor efficacy. Benefiting from the novel designs, \OURS{} surpasses all existing methods by a significant margin in terms of both accuracy and speed, even without the requirement of expensive re-training or rendering.
\end{abstract}
\vspace{-0.2cm}
\section{Introduction}
\label{sec:intro}
\begin{figure}
  \centering
  \includegraphics[width=\columnwidth]{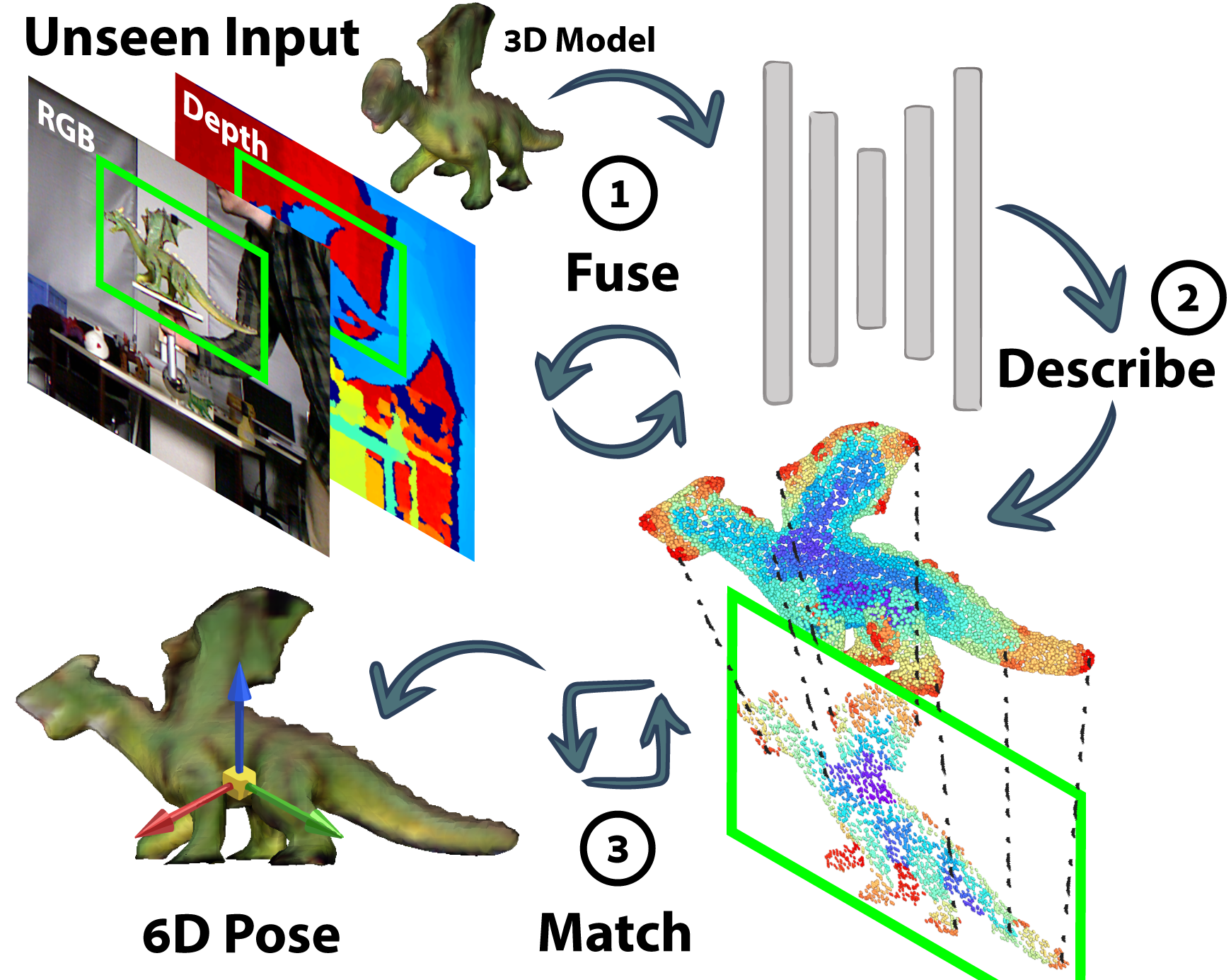}
  \vspace{-0.6cm}
  \caption{\OURS{} provides a pipeline to match a previously unseen 3D CAD model of an object to an RGBD image (Top left). (Fuse) Information from RGB-D and CAD is fused. (Describe) Consumes fused information and produces generic color-aware rotation-invariant 3D descriptors. (Match) Further used for establishing correspondences as well as the 6D pose.}  
  \label{teaser}
  \vspace{-0.6cm}
\end{figure}

Object 6D pose estimation is a critical task in computer vision applications, such as robotic manipulation~\cite{mason2018manip,zeng2016multi}, augmented reality~\cite{azuma1997AR,linowes2017augmented}, and autonomous driving~\cite{hoque2023driving,li2018stereo}. While object 6D pose estimation with object-specific training has achieved impressive results on benchmarks~\cite{sundermeyer2023bop}, handling unseen objects still remains a challenge. Approaches like template matching~\cite{kehl2017ssd}, keypoint detection~\cite{pvn3d,di2022gpv,wang2021gdr}, surface mapping~\cite{haugaard2022surfemb,su2022zebrapose}, and reconstruction-based frameworks~\cite{sun2022onepose,labbe2020cosypose,li2023nerf} have achieved high accuracy for individual objects~\cite{pvn3d,he2021ffb6d,haugaard2022surfemb,wang2021gdr,sun2022onepose}. However, these methods are not designed to handle multiple objects or generalize to objects not presented in the training data.

Dataset-level pose estimation methods~\cite{wang2019densefusion} can handle multiple objects in a dataset but struggle when faced with new instances. Similarly, category-level pose estimation methods~\cite{wang2019normalized} generalize to new instances within the same category but struggle with new categories. These approaches do not apply to the challenging problem of unseen object pose estimation in real-world applications where the 3D model is only available during inference time, since the models are designed to overfit the specific distribution of one object, category, or dataset. 
Some one-shot learning methods attempt to align object models using template matching or capture the structure from motion (SfM) of unseen objects \cite{osop, sun2022onepose}. However, these approaches often require object-specific preprocessing steps. Classic approaches that target unseen object pose estimation employ handcrafted features and correspondences between CAD models and observed RGB-D images \cite{drostppf, okorn2021zephyr, konig2020hybridppf}. However, these methods introduce many pose hypotheses with high ambiguity, which require to be rated and refined iteratively, resulting in computational overhead. 

Recent works have extensively delved into generic pose estimation through learning models that rate and refine pose hypotheses~\cite{busam2020like,megapose}, predominantly utilizing time-consuming render-and-compare strategies. This constraint limits their utility in real-world applications. Alternatively, some approaches formulate the generic 6D pose estimation problem as a point cloud registration task, benefiting from the point cloud representation backbones but neglecting crucial texture information from RGB images 
\cite{zeropose, gcpose}, leading to less distinct point cloud-based descriptors.  Therefore, they introduce ambiguities in correspondence extraction partially remedied by adding knowledge about the object symmetry during training \cite{gcpose}.

We present \OURS{}, a Fuse-Describe-Match strategy for unseen object 6D pose estimation from single RGB-D images as shown in Figure ~\ref{teaser}. 
Our method is designed to extract rotation-invariant descriptors that can be shared across a wide range of objects, facilitating generalization to unseen objects. The extraction of rotation-invariant descriptors is crucial as it allows our method to inherently capture and model the natural symmetry of objects without relying on explicit symmetry annotations.
However, rotation invariance still has some ambiguity where one point can be matched to several geometrically similar points. To address the ambiguity problem introduced by rotation invariance, we introduce a novel 2D-3D fusion module termed Latent Fusion Attention Module. This module effectively combines texture and geometric information. This results in extracting descriptors that describe both the appearance and the shape features of an object in a complementary and generic manner.
Furthermore, we propose a novel Bridged Coarse-level Matching loss that leverages RGB information to enhance the learning of geometric descriptors. This loss function strengthens the association between texture and geometric features, leading to more precise and accurate matching between CAD models and RGB-D images of unseen objects.
 Our main contributions are:
\begin{itemize}
  \item We propose \textbf{\OURS, a 6D pose estimation fuse-describe-match strategy} that extracts fused RGB-D input features targeted to register an unseen 3D CAD model to an object in the scene.
  \item We introduce a novel \textbf{Latent Fusion Attention Module} to effectively fuse texture and geometric features for generic pose estimation from RGB-D data and train \OURS~with a \textbf{Bridged Coarse-level Matching Loss}.
  \item \OURS~captures symmetries inherently by \textbf{learning a fused feature representation} without additional annotations thus reducing pose ambiguities.

\end{itemize}

\section{Related Work}
\label{sec:related}
The majority of related work focused on 6D pose estimation of seen objects for which training data (real or synthetic) is available. However, they need to be retrained for any new object instance. There were extensions to object category pose estimation, but they cannot generalize to new unseen categories. Therefore, in recent two years several approaches, that aim at generalizing to novel unseen objects without retraining, were introduced.

\subsection{Seen Object Pose Estimation.}
The approaches for seen object pose estimation rely on available real or synthetic training data and train one neural network model per object or per scene. They are usually multi-stage pipelines where the core learning efforts are in establishing image-to-model~(2D-3D) correspondences further used for pose estimation through PnP+RANSAC or direct regression. A larger amount of learning-based approaches consume RGB as input and only a few of them focus on RGB-D inputs facing the challenge of fusing RGB and depth information in neural networks.

\noindent\textbf{RGB-D Fusion Methods} are important because they profit from complementarity of two data sources and naturally improve pose accuracy as demonstrated in early works ~\cite{hinterstoisser2011multimodal, yu2012sparse}. In deep learning approaches, features are extracted separately from two modalities with different neural networks and their fusion is not obvious. Early approache, like PointFusion~\cite{xu2018pointfusion}, extracts global RGB (CNN) and depth (PointNet) features from the patch containing the object, and fuses them with per-point depth features for 3D object bounding box detection. Later, DenseFusion~\cite{wang2019densefusion} performs late per-point feature fusion strengthened with the global information, allowing better discrimination at the local level and resulting in better occlusion handling. Other works like PVNet3D~\cite{pvn3d} rely on DenseFusion and estimate sparse keypoints instead of dense correspondences. FFB6D~\cite{he2021ffb6d} instead uses bidirectional fusion modules to combine modality information at earlier stages and produce stronger per-pixel fused features.  Recently, DFTr~\cite{zhou2023DFTr} uses Transformers and improves the data fusion with the global semantic similarity between RGB and depth. This fusion strategy can help handle missing and noisy data caused by reflections or low-texture information.

\noindent\textbf{Symmetric Objects} are problematic because they look the same from different viewpoints. Correspondence-based methods have issues with visual ambiguities cause~\cite{manhardt2019explaining} as one-to-many matches define multiple equally correct poses. This has been tackled if symmetry information is known beforehand and used for data preparation~\cite{zakharov2019dpod} or in loss functions~\cite{wang2021gdr, gcpose}. Contrary to this SurfEmb~\cite{haugaard2022surfemb} does not require known symmetry and learns symmetry invariant features with contrastive loss. Learned 2D-3D descriptions from SurfEmb~\cite{haugaard2022surfemb} are not guaranteed to be invariant to rigid object transformations; it robustly learns quasi-invariance from a large dataset for specific objects. Additionally, predicting pose distribution~\cite{murphy2021implicit,iversen2022ki,haugaard2023spyropose} instead of a single estimate elegantly circumvents this problem.

\subsection{Unseen Object Pose Estimation}
Pose estimation of unseen objects considers that the neural network model is trained once and can generalize to novel unseen objects without retraining. For long, handcrafted feature matching using \textit{point pair features}(PPF)~\cite{drostppf} has been a competitive method in BOP challenge~\cite{sundermeyer2023bop} enabling unseen object pose estimation. Its main disadvantage is efficiency due to large voting spaces and adding RGB to PPF~\cite{drostppf} brought some benefits. Recently, Gen6D~\cite{liu2022gen6d}, OnePose~\cite{sun2022onepose} and OnePose++\cite{he2022oneposeplusplus}, utilize SfM and feature matching techniques to align a posed set of images of a given object to a target view using refined nearest neighbor image retrieval~\cite{liu2022gen6d} or 2D-3D image matching~\cite{sun2021loftr}.

Template-based methods like OSOP~\cite{osop} and OVE6D~\cite{cai2022ove6d} rely on representing the target object with templates. OSOP is a multi-stage pipeline, which leverages templates representing the 3D object seen from different views for segmentation, closest viewpoint selection and dense matching. OVE6D~\cite{cai2022ove6d} is inspired by early ideas to learn specific embedding spaces for pose estimation~\cite{wohlhart2015learning}, and, thus represents various 3D models together in an embedding space. 
ZePHyR~\cite{okorn2021zephyr} proposes hypothesis scoring, while MegaPose~\cite{megapose} proposes generic render-and-compare RGB pose refinement. The extracted features of OSOP~\cite{osop} are not invariant to rotations and the design of the ZePHyR~\cite{okorn2021zephyr} and MegaPose~\cite{megapose} are computation-intensive since they need to evaluate many sampled hypotheses.

A natural way towards the generalization to unseen objects is through descriptor learning, where generic descriptors can be used to match depth pixels of the object with its 3D model. Learning strong local 3D descriptors has been exhaustively studied in the context of point cloud registration~\cite{yu2021cofinet,huang2021predator,li2022lepard,Qin_2022_CVPR}. However, for object pose estimation descriptors need not only to be unique and repeatable, but also require rotation invariance introduced in RIGA~\cite{yu2022riga}, YOHO~\cite{wang2021you}, and RoITr~\cite{yu2023roitr}. Moreover, adding color information to them and maintaining generality is an additional challenge. Recent works tackling unseen object pose estimation, like Zeropose~\cite{zeropose} and GCPose~\cite{gcpose} rely on 3D descriptor learning. Zeropose~\cite{zeropose} uses the foundation models  ImageBind~\cite{girdhar2023imagebind} and SAM~\cite{kirillov2023segment} together with 3D-3D feature matching using descriptions from GeoTransformer~\cite{Qin_2022_CVPR}.
GCPose~\cite{gcpose} uses the same descriptor principle~\cite{Qin_2022_CVPR} but with explicit knowledge of object symmetries. 

We leverage RoITr's~\cite{yu2023roitr} rotation-invariance by design and do not require pre-defined object symmetries for training.  We aim to get the best of both worlds by designing a rotation-invariant and symmetry-aware backbone, which fuses RGB and depth information efficiently.

\begin{figure*}[t]
  \centering
  \includegraphics[width=\textwidth]{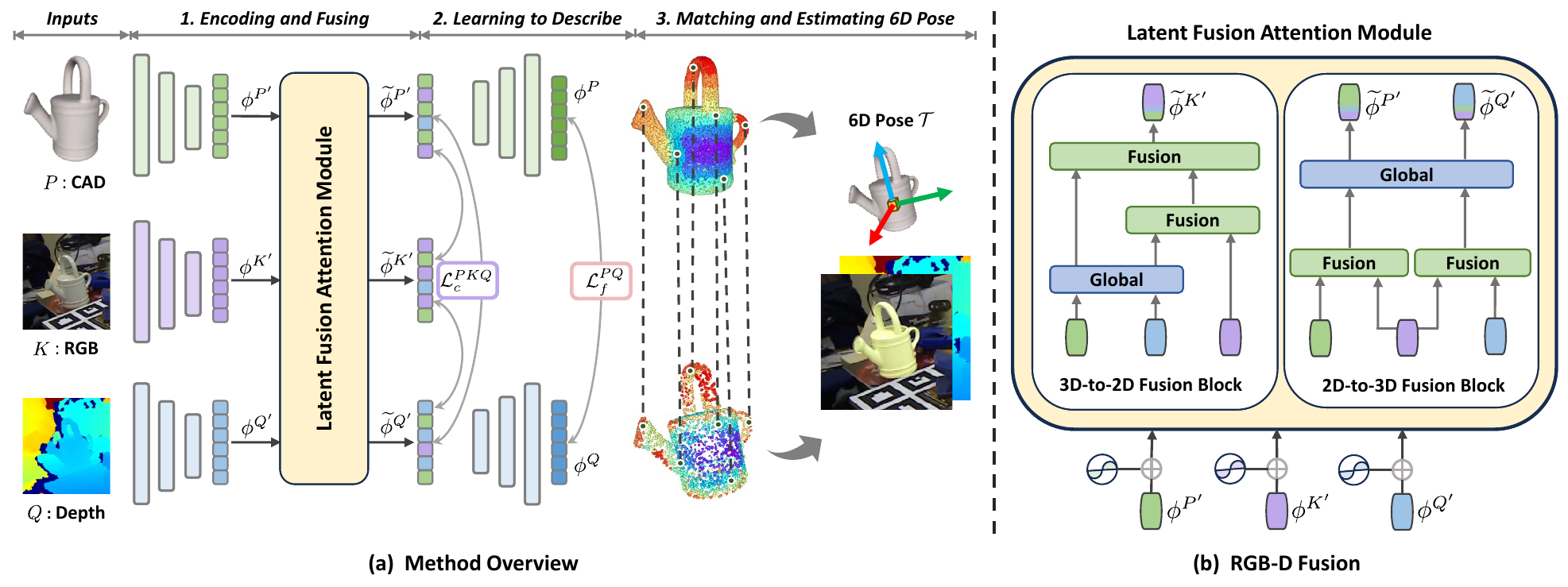}
  \caption{\textbf{Overview of \OURS{}.} 
  Upon encountering an unseen object, we initially derive the segmented depth point cloud $Q$ and the corresponding RGB image crop $K$ utilizing a pre-trained generic segmentation network. Subsequently, we procure both 3D and 2D local features from the CAD point cloud $P$, depth point cloud $Q$, and the RGB image crop $K$. These extracted features are then amalgamated within a latent space through our innovative \textit{Latent Fusion Attention Module}, under the guidance of a \textit{Bridged Coarse-level Matching Loss (BCM Loss)} $\mathcal{L}{c}^{PKQ}$. The refined 3D descriptors $\widetilde{\phi}^{P'}$ and $\widetilde{\phi}^{Q'}$ are fed into decoders, which enhance the resolution of the descriptors to ${\phi}^{P}$ and ${\phi}^{Q}$, this process being steered by a detailed matching loss $\mathcal{L}^{PQ}_{f}$. In the final stage, the 6D pose of the novel objects is deduced by aligning the descriptors within the latent space and aggregating the pose parameters $\mathcal{T}$.
  }
  \label{fig:overview}
  \vspace{-0.5cm}
\end{figure*}



\section{Method}
\label{sec:method}

\subsection{Problem Formulation}

The task of unseen object pose estimation aims at estimating the 6D pose between a CAD model, which is not available during training, and its partial observation from the RGB and/or depth image. In this paper, we estimate the pose of unseen objects by matching the learned descriptors between RGB-D data and its CAD model. The input of our method includes an unseen CAD model represented as a point cloud $P=\{p_i\in \mathbb{R}^3 \mid 1 \leq i \leq n\}$ with $n$ points, the partial point cloud obtained from the depth channel, denoted as $Q=\{q_j\in \mathbb{R}^3 \mid 1 \leq j \leq m\}$ with $m$ points, as well as its corresponding RGB image crop $K$ of the localized object. 
Corresponding points $p_{i} \leftrightarrow q_{j}$ are collected in the predicted correspondence point set $\mathcal{C}$ which is used to estimate the 6D pose of the novel object, by optimizing the objective
\begin{equation}
  \label{eq:optimization}
  \begin{aligned}
  \mathcal{T^{*}} = \arg\min_{\mathcal{T} \in \text{SE}(3) } \sum_{(p_{i}, q_{j}) \in \mathcal{C}} \|(\mathcal{T}p_{i} - q_{j})\|^{2}_{2}
\end{aligned}
\end{equation}
of mutual 3D correspondences. $\mathcal{T} \in \text{SE}(3)$ denotes the 6D pose of the novel object in the Special Euclidean group $\text{SE}(3)$ of rigid transformation in 3D space.

\subsection{Method Overview}

To solve for object pose, we first calculate correspondences by extracting the generic descriptors $\mathcal{\phi}^{P}$ and $\mathcal{\phi}^{Q}$ for the points in $P$ and $Q$ in latent space $\mathbb{R}^d$.
By calculating the similarity between $\mathcal{\phi}^{P}$ and $\mathcal{\phi}^{Q}$, we can construct the correspondence set as $\mathcal{C}= \{(p_{i}, q_{j}) \mid \mathcal{\phi}^{P}_{i} \leftrightarrow \mathcal{\phi}^{Q}_{j}\}$,
where $\mathcal{\phi}^{P}_{i} \leftrightarrow \mathcal{\phi}^{Q}_{j}$ denotes matched descriptors of the point $p_{i}$ and $q_{j}$, respectively.
The optimization problem (Eqn.~\ref{eq:optimization}) is designed as a least square problem that can be robustly solved with an outlier-aware consensus algorithm such as RANSAC~\cite{ransac}.

 From the input CAD point cloud $P$, depth point cloud $Q$, and the RGB image $K$, our proposed method estimates a mapping function $\mathcal{\psi}$ that maps $P$ and $Q$ to generic descriptors $\mathcal{\phi}^{P}=\mathcal{\psi}(P\mid(Q,K)) \in \mathbb{R}^{n \times d}$ and $\mathcal{\phi}^{Q}=\mathcal{\psi}(Q\mid(P,K)) \in \mathbb{R}^{m \times d}$ by fusing the cross-modality information from $(Q, K)$ and $(P, K)$, respectively. By matching our learned generic descriptors, correspondences are established between the unseen object and its partial observation, and the object pose is finally estimated. An overview of our framework is depicted in Fig.~\ref{fig:overview}.

\subsection{Encoding and Fusing Descriptors}

We first introduce the extraction of 3D and 2D local features, and then the cross-modality descriptor fusion.

\noindent\textbf{Local 3D Feature Extraction.} We employ the recent transformer-based architecture RoITr~\cite{yu2023roitr} as our encoder backbone to extract rotation-invariant 3D local features from the CAD point cloud $P$ and the partially observed point cloud $Q$ from depth image.
The inherent rotation-invariance of the descriptor provides a robust feature extraction for geometric cues and guarantees the generalizability for unseen objects.
Given $P$ and $Q$, our encoder down-samples the input point clouds via Farthest Point Sampling~(FPS) to superpoints.
They represent a well-distributed coarse representation of spatial structure from the underlying dense point cloud defined as $P^\prime= \{p^\prime_i\in \mathbb{R}^3 \mid 1 \leq i \leq n^\prime\}$ and $Q^\prime= \{q^\prime_j\in \mathbb{R}^3 \mid 1 \leq j \leq m^\prime\}$, 
where $n^\prime$ and $m^\prime$ stand for the number of superpoints in $P^\prime$ and $Q^\prime$, respectively. Following~\cite{yu2023roitr}, for each superpoint $p^\prime_i$ and $q^\prime_j$, we first extract the local geometric features from neighboring points within a radius of $r$.
The local geometric cues are then projected into the latent space by a sequence of attention blocks, from which we obtain the inherently rotation-invariant local 3D geometric descriptors, denoted as $\phi_{p'_i} \in \mathbb{R}^d$, and $\phi_{q'_j} \in \mathbb{R}^d$ where $d$ is the dimension of the latent space. 

\noindent\textbf{Local 2D Feature Extraction.} A convolutional neural network (CNN) is used for local visual feature extraction.
Following LoFTR~\cite{sun2021loftr}, we adopt a modified encoder of FPN~\cite{fpn} as our CNN backbone. This 2D encoder down-samples the input image crop of size $H \times W$ to a feature map of size  $\frac{H}{8} \times \frac{W}{8}$, while simultaneously projecting the local textural information into a $d$-dimension latent space consistent with the 3D geometry features.
The image's local feature map is then flattened into $\phi^{K'}= \{k_t\in \mathbb{R}^d\big| 1 \leq t \leq \frac{H}{8} \times \frac{W}{8}\}$, where we denote the 2D superpixels as $K'$ and the 2D superpixel features as $\phi^{K'}$.


\noindent\textbf{Latent Fusion Attention Module.} After extracting 3D and 2D local features, we fuse the encoded 3D and 2D context in latent space via our proposed Latent Fusion Attention Module.
To keep the generalizability of our network and avoid overfitting on object-specific features, we propose to fuse the 3D and 2D features in a coarse-level latent space and leverage a \textit{3D-to-2D Fusion Block} as well as a \textit{2D-to-3D Fusion Block} for fusing the information in two perspectives. We leverage the \textit{Latent Fusion Transformers}~(green layers) and \textit{Global Transformers}~(blue layers) in these two fusion blocks as shown in Figure ~\ref{fig:overview}~(b).

Previous methods~\cite{he2021ffb6d,pvn3d,wang2019densefusion} usually interpolating features w.r.t. their spatial relationships explicitly. In MatchU, we use the positional encoding in the attention mechanism to incorporate spatial awareness and implicitly align different modalities. For 2D features, we follow DETR~\cite{detr} to encode the spatial information of the 2D feature map into the feature space. As for 3D, instead of encoding the raw position of the points~\cite{zhou2023DFTr}, we propose to use the pose-agnostic Point Pair Features~(PPFs)~\cite{drostppf} as the position representation following ~\cite{yu2023roitr}, which guarantees the geometric rotation-invariance and generalizability for unseen object pose estimation.

\noindent The \textit{Latent Fusion Transformer} is designed to fuse the 2D superpixel features and 3D superpoint features in the latent space, which consists of a series of self-attention and cross-attention layers. Following~\cite{sun2021loftr}, we adopt the linear attention ~\cite{katharopoulos2020transformers} for all the self- and cross attention layers 
with the goal of lower computational complexity.
We stack $g$ self- and cross-attention layers for each Latent Fusion transformer in practice.
The \textit{Global Transformer} is designed to aggregate the global context of the 3D and 3D features, for which we follow the design of RoITr~\cite{yu2023roitr}.

Details for both \textit{3D-to-2D Fusion Block} and \textit{2D-to-3D Fusion Block} are illustrated in Fig.~\ref{fig:overview}.
For the \textit{3D-to-2D Fusion Block}, we first aggregate CAD $\phi^{P'}$ and depth $\phi^{Q'}$ superpoint features with a \textit{Global Transformer}.
Then the RGB feature $\phi^{K'}$ is fused with the global-aware depth and CAD feature sequentially to get the final cross-modal 2D feature $\widetilde{\phi}^{K'}$ for each superpixel by \textit{Latent Fusion Transformer}.
For the \textit{2D-to-3D Fusion Block}, we first separately enhance both CAD $\phi^{P'}$ and depth $\phi^{Q'}$ superpoint features with RGB features through \textit{Latent Fusion Transformer}. A \textit{Global Transformer} then co-injects this information to provide the 2D-aware 3D superpoint features $\widetilde{\phi}^{P'}$ and $\widetilde{\phi}^{Q'}$ for both CAD and depth.

\subsection{Learning to Describe}
In order to guide the learning of the fused descriptors, we propose several loss functions. With the latent features learned from RGB images as the bridge between the latent spaces of the CAD and depth point clouds, we define \textit{Bridged Coarse-level Matching Loss}~(BCM Loss), which significantly facilitates the unification of two different 3D-based latent spaces, and helps to generate more robust and reliable correspondences between superpoints. Moreover, a fine-level matching loss is also introduced to guide the refinement of superpoint matches to point correspondences.

\noindent\textbf{Bridged Coarse-level Matching Loss.} To ensure the effectiveness of RGB-based 2D information in the latent space, the key is to provide the supervision signal from both 2D and 3D modality by establishing the cross-modal matches between 2D and 3D features. The alignment between the superpoints $P'$ and $Q'$ can be obtained via ground-truth transformation matrix and nearest neighbor search. Ground-truth 3D-2D correspondences between superpoints and superpixels are inherent in the RGB-D pair.

 We adopt the Circle Loss~\cite{circle}, which maximizes the similarity of the positive pairs of the superpoints, as well as minimizes the similarity of the negative pairs of the superpoints. Specifically, for each superpoint $p'_{i}\in P'$, and $q'_{j}\in Q'$, we can calculate the overlap $\mathcal{V}$ between $p'_{i}$ and $q'_{j}$ as:
\begin{equation}
  \label{eq:overlap}
    \mathcal{V}(p'_{i}, q'_{j}) = \frac{|\{\hat{p'}_{u}\in\hat{P'}_{i} \mid \exists\ \hat{q'}_{v}\in \hat{Q'}_{j} 
    : \hat{p'}_{u} \leftrightarrow \hat{q'}_{v}\}|}{|\{\hat{p'}_{u} \in \hat{P'}_{i}\}|},
\end{equation}

\noindent where $\leftrightarrow$ denotes the correspondence relationship. $\hat{P'}_{i}$ is the group of points from $P'$ assigned to $p'_{i}$ by Point-to-Node grouping strategy~\cite{yu2021cofinet}, and $\hat{Q'}_{j}$ means the same for $Q^\prime$.
A pair of superpoints $p'_{i}$ and $q'_{j}$ are considered as a positive pair if and only if
$\mathcal{V}(p'_{i}, q'_{j}) > \tau_r$, where $\tau_r$ is the
threshold of the overlap.
We sample a positive set of superpoints from $Q'$, and a negative set of superpoints for ${P}^\prime$. The coarse-level superpoint Circle Loss loss for $P'$ can be calculated with the weight of overlap, which we denote as $\mathcal{L}^{P'}_{c}$. We provide the detailed loss function in Appendix.

The same loss for $Q'$ is defined similarly, 
and the overall loss between the superpoints $P'$ and $Q'$ is defined as
\begin{align}
    \mathcal{L}_{c}^{P'Q'} = (\mathcal{L}^{P'}_c + \mathcal{L}^{Q'}_c)/2.
\end{align}
Similarly, we apply Circle Loss for 3D-2D coarse-level matching. We first project the 3D positive and negative samples into 2D plane, and then obtain the positive and negative pairs between the 2D superpixels and 3D superpoints. 

The loss function between $K'$ and $Q'$ is defined as $\mathcal{L}^{K'Q'}_{c}$,
and the loss function between $K'$ and $P'$ is defined as $\mathcal{L}^{K'P'}_{c}$.
Then we calculate the overall Bridged Coarse-level Matching Loss between $P'$, $Q'$ and $K'$ as
\begin{equation}
  \label{eq:coarse}
    \mathcal{L}_{c}^{PKQ} = \lambda_{b}\mathcal{L}_{c}^{P'Q'} + (1-\lambda_{b})(\mathcal{L}_{c}^{K'Q'} + \mathcal{L}_{c}^{K'P'}),
\end{equation}
where $\lambda_{b}$ and 1-$\lambda_{b}$ is the weight for the 3D-3D and 3D-2D matching loss respectively.

\noindent\textbf{Fine-level Matching Loss.} In order to enhance the precision of the 3D-3D correspondence, we apply a fine-level matching loss to the CAD 
point cloud $P$ and the observed point cloud $Q$.
We use a series of decoder blocks introduced in \cite{yu2023roitr}, which generates denser points $P$ and $Q$ from the coarse-level superpoints $P'$ and $Q'$. 
Given the superpoint correspondence, the group of fine-level point features is assigned to each superpoint through the point-to-node strategy, and the similarity matrix can be calculated between the corresponding groups. The fine-level matching is formulated as 
an optimal transport problem, which can be solved by the Sinkhorn algorithm ~\cite{sinkon}. A negative log-likelihood is applied to the similarity matrix to obtain the fine-level matching loss $\mathcal{L}_{f}^{PQ}$ between $P$ and $Q$.
The overall loss function for our training is defined as:
\begin{equation}
  \label{eq:overallloss}
    \mathcal{L} = \lambda_{c}\mathcal{L}_{c}^{PKQ} + (1-\lambda_{c})\mathcal{L}_{f}^{PQ},
\end{equation}
where $\mathcal{L}_{c}^{PKQ}$ is the Bridged Coarse-level Matching Loss and $\mathcal{L}_{f}^{PQ}$ is the fine-level matching loss. $\lambda_{c}$ is the weight to balance the coarse and fine-level training.

\begin{table*}[t]
    \centering
    \begin{tabular}
    {@{\hspace{2pt}}l@{\hspace{5pt}}l@{\hspace{10pt}}c@{\hspace{10pt}}c@{\hspace{10pt}}c@{\hspace{5pt}}c@{\hspace{5pt}}c@{\hspace{5pt}}c@{\hspace{5pt}}c@{\hspace{5pt}}c@{\hspace{5pt}}c@{\hspace{5pt}}c@{\hspace{5pt}}c@{\hspace{2pt}}}
    
    \toprule[1.1pt]

    \multirow{2}{*}{} & \multirow{2}{*}{Method}& \multirow{2}{*}{MH} & \multicolumn{2}{l}{Obj. Loc.}  & \multirow{2}{*}{Refine.} & \multirow{2}{*}{LM-O} & \multirow{2}{*}{T-LESS} & \multirow{2}{*}{TUD-L} & \multirow{2}{*}{IC-BIN} & \multirow{2}{*}{YCB-V} & \multirow{2}{*}{Mean} & \multirow{2}{*}{Time(s)} \\
    \multicolumn{2}{c}{}&  &unseen &seen &  &  & & & & & \\ \hline
    \multirow{5}{*}{\begin{tabular}[l]{@{}l@{}}\textit{(a) one hypo.}\\\textit{w/o refine.} \end{tabular}} 
    & ZeroPose  &  & \ding{51} &  &  & 26.0  & 17.8 &  41.2 & 17.7  & 25.7  & 25.7  & 0.30  \\
    &OSOP  &   & \ding{51} &  &  & 39.3  & -  &-  & -  &\textbf{52.9}  &46.1  & 0.47      \\
    &ZeroPose &    &  & \ding{51}  &  & 26.1  & 24.3  & 61.1 & 24.7 & 29.5 & 33.1 & 0.30 \\
    &MegaPose &    &  & \ding{51}  &  & 18.7  & 19.7  & 20.5 & 15.3 & 13.9 & 17.6 & 2.50  \\
    &Ours(Fast)  &  & \ding{51} & & & \textbf{52.6} & \textbf{42.9}  & \textbf{70.0} & \textbf{36.7} & 50.5 & \textbf{50.5} & 0.07 \\
    \hline
    \multirow{2}{*}{\begin{tabular}[l]{@{}l@{}}\textit{(b) multi-hypo.}\\\textit{w/o refine.} \end{tabular}} 
    &OSOP  & \ding{51}  & \ding{51} &  &  & 46.2  & - &- & - & 54.2 &50.2  & 5.30 \\
    & Ours(Accurate) & \ding{51} & \ding{51} &  & &  \textbf{56.2} & \textbf{50.6} & \textbf{75.6} &\textbf{42.2} &\textbf{60.8} &  \textbf{57.8}  & 1.03 \\ 
    \hline
    \multirow{5}{*}{\begin{tabular}[l]{@{}l@{}} \textit{(c) unseen loc},\\\textit{w/ refine.}\\ \end{tabular}} 
    & DrostPPF  &\ding{51}   & \ding{51} &  & \ding{51} &52.7 & - & - & - & 34.4 & 43.6 & 15.9  \\
    &PPF + Zephyr &\ding{51}  & \ding{51}  &  & \ding{51} & 59.8 & - & -  & -  & 51.6 & 55.7  & 2.90 \\
    &OSOP   & \ding{51}  & \ding{51}  &  & \ding{51}  & 48.2 & - & - & - & 57.2 & 52.7  & 5.44 \\
    &ZeroPose  & \ding{51} & \ding{51} & & \ding{51}  & 49.1 & 34.0 & 74.5 & 39.0 & 57.7 & 50.9 & 6.75 \\
    &ZeroPose~  (BOP)  & \ding{51} & \ding{51} & & \ding{51}  & 53.8 & 40.0 & 83.5 & 39.2 & 65.3 & 56.4 & 6.75 \\
    &Megapose (BOP)  & \ding{51} & \ding{51} & & \ding{51}  & 62.6 & 48.7 & 85.1 & \textbf{46.7} & \textbf{76.4} & 63.9 & 10.70 \\
    &Ours(Accurate) & \ding{51} & \ding{51} & &\ding{51} &\textbf{64.4} & \textbf{52.7} & \textbf{89.8} & 44.2 & 72.6 & \textbf{64.7} &5.60\\
    \hline
    
    \multirow{3}{*}{\begin{tabular}[l]{@{}l@{}} \textit{(d) seen loc.},\\\textit{w/o refine.}\\ \end{tabular}} 
    &OVE6D  & \ding{51} &    &  \ding{51} &  & 49.6  & 52.3  & - & - & - & 57.5 & - \\
    &GCPose & \ding{51}  &  & \ding{51} &  & 65.2  & \textbf{67.9}  &92.6  &-  &-  & \textbf{75.2}  &  -  \\
    &Ours(Accurate) & \ding{51} & & \ding{51} & &\textbf{66.8} & 65.1 & \textbf{93.1} & 43.9 & 65.8 & 66.9 &3.12\\
    \hline
    \multirow{5}{*}{\begin{tabular}[l]{@{}l@{}} \textit{(e) seen loc.},\\ \textit{w/ refine.}\\ \end{tabular}}
    & OVE6D  & \ding{51}      & &\ding{51} & \ding{51}  & 62.7  & 54.6   & -     & -   & -  & 58.7   & - \\
    &HybridPPF  & \ding{51}  & &\ding{51} & \ding{51}  & 63.1  & 65.5   & 92.0  & -   & -  & \textbf{73.5}  & -  \\
    &Megapose   & \ding{51}  & &\ding{51} & \ding{51}  & 58.3  & 54.3   & 71.2  & 37.1 & 63.3  & 56.8   & 10.70  \\
    &ZeroPose   & \ding{51}  & &\ding{51} & \ding{51}  & 56.2  & 53.3   & 87.2  & 41.8 & 58.4 & 59.4  & 6.75                    \\
    &Ours(Accurate) & \ding{51} & & \ding{51} & \ding{51} &\textbf{68.0} & \textbf{66.8} & \textbf{94.7} & \textbf{47.4} & \textbf{75.6} & \textbf{70.5}  &5.60\\ 
    
    \bottomrule[1.1pt]

    \end{tabular}
    \captionsetup{width=\textwidth} 
    \caption{ \textbf{Quantitative results \protect \footnotemark in terms of Average Recall(AR) on BOP-5 core benchmark datasets for unseen object pose estimation task.}
    \textbf{MH}: whether multiple hypotheses were adopted. \textbf{Obj. Loc.}: whether the object localization (detection or segmentation) is trained on the test objects. \textbf{Refine.}: whether the result is refined with either depth and/or RGB images.
    }
    \vspace{-0.5cm}
    \label{tbl:main table}
    \end{table*}

\subsection{Matching Descriptors and Estimating 6D Poses}
\label{sec:6dpose}

After training on a large number of objects and images, we obtain a robust descriptor.
During inference, we utilize extracted features to establish 3D-3D matches between CAD point cloud $P$ and observation $Q$.
Following \cite{yu2023roitr}, we measure similarities of normalized features using covariance analysis.
We determine the top $\kappa$ most correlated ones as putative 3D-3D matches of the correspondence point set $\mathcal{C}$ from which we create $\eta$ pose hypotheses.
For each hypothesis $\mathcal{T}_v$ with $1 \leq v \leq \eta$, we first randomly select $s \ll \kappa$ correspondences from $\mathcal{C}$ and then solve Eq.~\ref{eq:optimization} for the 6D object pose using RANSAC~\cite{ransac} optimization.
This process speeds up the prediction process and provides us with the control parameter $\eta$ to determine the efficiency of \OURS.
All hypotheses are then ranked by an average score between 3D and RGB verification processes.
For 3D, we calculate the score based on the Euclidean point-to-point distance between the transformed CAD model and the lifted depth map. For RGB, we follow the proposal of~\cite{megapose}.
Our final prediction is the pose $\mathcal{T}$ with the highest score.

\vspace{-0.3cm}
\section{Experiments and Results}
\label{sec:experiments_and_results}
\subsection{Implementation Details}
During training, default settings are $\lambda_b=0.3$ and $\lambda_c=0.5$.  For inference, we set $\kappa=128$, $s=64$, and initially $\eta=20$, adjusting $\eta$ to 64 to enhance accuracy during testing. To localize unseen objects in RGB-D images, we utilize a CAD-based segmentation method~\cite{nguyen2023cnos}. For 6D pose evaluation with instance-level localization, existing detection results from~\cite{zeropose,Li_2019_ICCV,Sundermeyer_2020_CVPR} are employed. 
Detailed implementation and more information about the network designs are provided in the Appendix.

\footnotetext{ The numbers and timings are either from the original papers~\cite{zeropose,osop} or BOP Challange~\cite{sundermeyer2023bop}.
} 

\subsection{Datasets and Evaluation Metrics}
\paragraph{Datasets.} Following ZeroPose \cite{zeropose} and MegaPose \cite{megapose}, we utilize the Google-Scanned-Objects (GSO) dataset \cite{gso} provided by MegaPose to train our model, where 850 GSO objects with around 800K rendered images are used for training and the rest 94 objects with around 200K images are for validation. To evaluate our method on unseen object pose estimation, we employ five core BOP datasets as our testing set, \ie LM-O~\cite{brachmann2014learning},T-LESS~\cite{hodan2017tless}, TUD-L~\cite{hodan2018bop}, IC-BIN~\cite{doumanoglou2016recovering}, and YCB-V~\cite{xiang2018posecnn}, among which the LM-O dataset is adopted for our ablation studies. All the CAD models and images in the test set are guaranteed to be unseen during training.
 %

\noindent\textbf{Metrics.} 
 We adopt the Average Recall~(AR) in the standard benchmark BOP~\cite{hodan2018bop,sundermeyer2023bop} as our main evaluation metric. It calculates the average recall of three pose errors by varying the thresholds in a determined range.
We also adopt the average distance metric (ADD) as our secondary metric for a fair comparison with baselines in Table \ref{tbl:compare to ffb}. ADD calculates the average point distance between the point clouds of the object CAD model with ground-truth and estimated pose. We report the accuracy of distance less than 10\% of the objects' diameter (ADD-0.1d) as ~\cite{he2021ffb6d,zhou2023DFTr,pvn3d}.

\begin{figure}[h]
  \centering
 \includegraphics[width=\columnwidth]{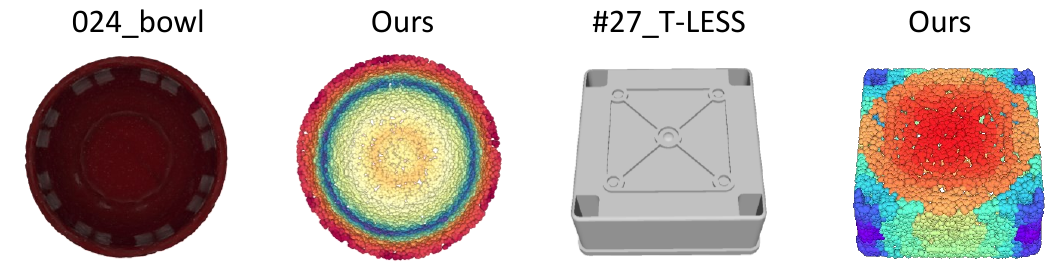}
  \caption{\textbf{t-SNE visualization of our descriptors for symmetric objects.}  We showcase the capability of capturing both continuous and discrete symmetries without external annotation.}
  \vspace{-0.5cm}
  \label{fig:symmetry vis}
\end{figure}

\begin{figure}[h]
  \centering
 \includegraphics[width=\columnwidth]{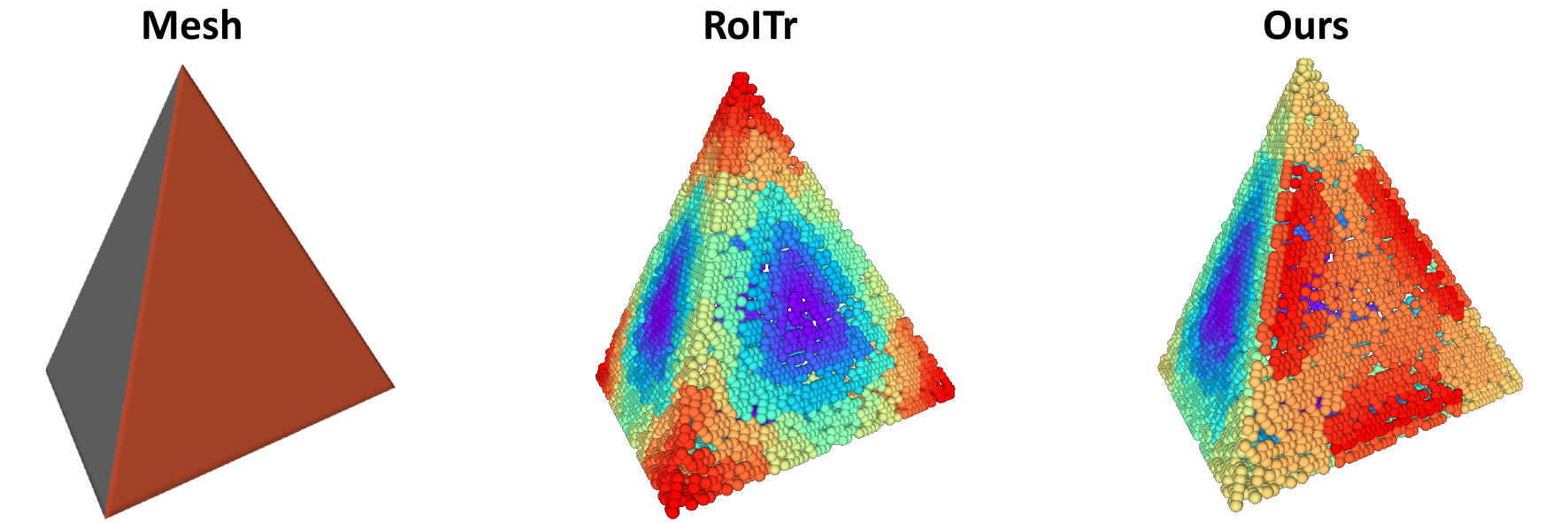}
  \caption{\textbf{t-SNE visualization of a regular tetrahedron with different colored faces} (left). Our method can extract distinct descriptors for the red face (right) while RoITr cannot (center).}
  \vspace{-0.5cm}
  \label{fig:tetX}
\end{figure}

\subsection{Evaluation on 5 BOP Core Benchmark Datasets}
\label{bop benchmark}
\begin{figure*}[t]
  \centering
 \includegraphics[width=\textwidth]{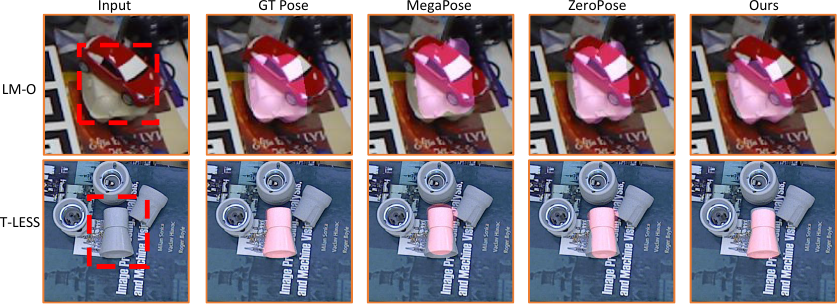}
\vspace{-0.6cm}
  \caption{\textbf{Qualitative results of 6D pose estimation of our method in comparison with Megapose and ZeroPose.} 
  The upper row shows an egg box which is heavily occluded in the LM-O dataset. Our method is robust to handle occlusion while other methods flip the poses by mistake. The lower row shows a highly ambiguous object that other methods put the pose upside down but ours predicts accurately. 
  }
  \vspace{-0.3cm}
  \label{fig:posevis}
\end{figure*}

The evaluation results of our method are shown in Tab.~\ref{tbl:main table}.  Compared with all the baselines, we achieve state-of-the-art performance on the task of unseen object pose estimation. We denote our method using only 1 pose hypothesis as Ours~(Fast)  and as Ours~(Accurate) when multiple hypotheses are leveraged. For a fair comparison, we run our method under 5 specific settings in accordance with the baselines.
In \textit{(a)}, we compare our method with 
ZeroPose~\cite{zeropose}, OSOP~\cite{osop}, and MegaPose~\cite{megapose} with only one hypothesis during the inference stage and without any refinement. Under \textit{(a)}, Ours~(Fast) performs significantly better and faster on average. Note that in \textit{(a)}, we use a generic instance detection/segmentation network for unseen objects~\cite{nguyen2023cnos}.
This demonstrates the robustness of our method against noisy detection initialization, which often occurs in real applications.
In \textit{(b)}, we further improve our results by introducing more pose hypotheses~(20 by default) as we described
in Section~\ref{sec:6dpose}. We compare our method with OSOP~\cite{osop} with the same number of hypotheses and exhibit superior performance. 
In  \textit{(c)}, by adding the ICP refinement, our model achieves the best results
among all the methods on both the overall and the per-dataset evaluation. Notably, in ~\textit{(c)}, our method requires less time to perform one inference compared to other baselines~\cite{zeropose,megapose} with rendering-based refinement. 
In  \textit{(d)}, we compare our method with OVE6D~\cite{cai2022ove6d} and GCPose~\cite{gcpose} with an identical object detection/segmentation network. Our method consistently surpasses the baselines on T-LESS and TUD-L datasets.
In  \textit{(e)}, we finally improve our results by using a trained detector and incorporating ICP refinement. We showcase the qualitative results in Figure~\ref{fig:posevis}. 

\begin{table}[t]
\centering
\begin{tabular}{@{\hspace{5pt}}l@{\hspace{10pt}}c@{\hspace{10pt}}c@{\hspace{10pt}}c@{\hspace{10pt}}c@{\hspace{5pt}}}
   \toprule[1.1pt]
    Method & Object Loc. & Pose & ADD-0.1d\\ \hline
    PVN3D~\cite{pvn3d}  & seen  & seen   & 63.2 \\
    FFB6D~\cite{he2021ffb6d}  & seen  & seen   & 66.2 \\
    DFTr~\cite{zhou2023DFTr}   & seen  & seen   & \textbf{77.7} \\
    Ours(1hypo)    & seen   & seen & 68.4 \\
    Ours(20hypo)   & seen   & seen &  75.7 \\
    \hline
    Ours(1hypo)    & unseen & unseen & 61.7 \\
    Ours(20hypo)   & unseen & unseen & \textbf{70.8}\\

\bottomrule[1.1pt]
\end{tabular}
\caption{\textbf{Quantitative evaluation of 6D pose (ADD-0.1d) on
the LM-O dataset for seen object pose estimation task.}}
\label{tbl:compare to ffb}
\vspace{-0.3cm}
\end{table}
\subsection{Capturing Symmetry and Texture}
\label{vis symmetry}
We visualize our descriptors for symmetric objects with t-SNE~\cite{van2008visualizing}. As shown in Figure ~\ref{fig:symmetry vis}, our learned descriptors can capture continuous~(024\_bowl) and discrete~(\#27\_T-LESS) symmetries, which is credited to the rotation-invariant property of our design. Compared with GCPose~\cite{gcpose}, which relies on the supervision of symmetry labels, our descriptors recognize the symmetry even without any external symmetry annotations. Moreover, we visualize ``tetX'' from SYMSOL~\cite{murphy2021implicit} dataset, a regular tetrahedron with one red and three white faces in Figure~\ref{fig:tetX}. The descriptors extracted by RoITr show the same distribution on all 4 faces due to their geometric similarity, introducing ambiguity issues for matching and leading to incorrect pose estimation potentially. In contrast, our method extracts distinct descriptors on the textured face. This indicates that our method not only describes the geometric property of the objects but also captures the texture information which further eliminates ambiguities in pose estimation.

\subsection{Comparison with RGB-D Fusion Pipelines}
\label{compare with rgbd fusion methods}
To demonstrate the effectiveness of our proposed RGB-D fusion mechanism, we compare \OURS{} with recent RGB-D fusion approaches ~\cite{wang2019densefusion,he2021ffb6d,pvn3d,zhou2023DFTr} 
 on LM-O dataset. 
As shown in Tab.~\ref{tbl:compare to ffb}, although our method is specifically designed for unseen object pose estimation, it still outperforms PVN3D~\cite{pvn3d} as well as FFB6D~\cite{he2021ffb6d}, and achieves comparable results with the current state-of-the-art method DFTr~\cite{zhou2023DFTr} by increasing the number of hypotheses, this demonstrates the efficacy of our RGB-D fusion mechanism. 
Moreover, our method trained without the test objects even outperforms most of the baselines that have seen them. This result further confirms the generalizability of our method. 

\subsection{Ablation Study}
\label{ablation}
\paragraph{Key Design Principles.}
First, we
replace the BCM Loss $\mathcal{L}_{c}^{PKQ}$ with only a 3D coarse matching loss $\mathcal{L}_{c}^{P'Q'}$. The performance drops obviously as shown in the first row of Table~\ref{tbl:compare to roitr}, proving the effectiveness of BCM Loss in guiding the descriptor learning. Second, we mask out the input RGB image. A sharp decrease in performance (2nd row) indicates the efficacy of our RGB-D fusion, as well as
the pivotal role that RGB information plays in our pipeline. 
Third, we take the original RoITr model and initialize the point features with the RGB values to further demonstrate the superiority of our RGB-D fusion. RoITr \textit{w/} RGB surpasses Ours \textit{w/o} RGB, but is still inferior to our full pipeline by a large margin.

\begin{table}[t]
  \vspace{-0.2cm}
  \centering
  \begin{tabular}{lc}
    \toprule[1.1pt]
    Method    &Mean AR~(BOP-5)\\ \hline
     Ours ~\textit{w/o} ~BCM Loss    &48.1 \\
     Ours ~\textit{w/o} ~RGB Input   &39.6 \\
     RoITr ~\textit{w/} ~RGB Init.  &42.6 \\
     Ours & \textbf{50.5} \\ 
     \bottomrule[1.1pt]
 \end{tabular}
  \caption{\textbf{Ablation study of our key designs on BOP-5 datasets}. 
  }
\vspace{-0.5cm}
\label{tbl:compare to roitr}
\end{table}

\begin{figure}[t]
  \centering
  \vspace{-0.2cm}
 \includegraphics[width=\columnwidth]{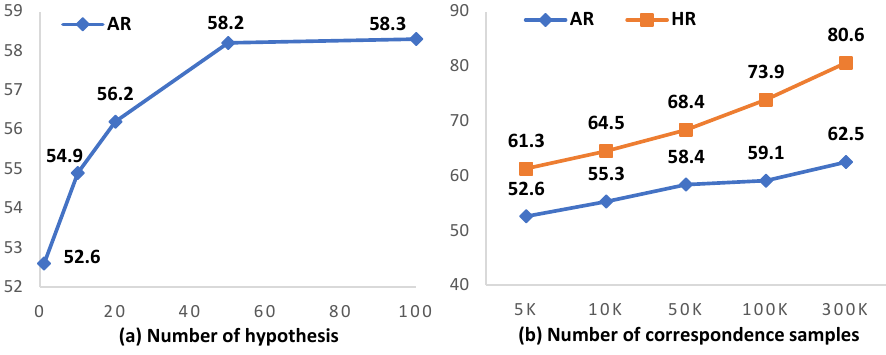}
 \vspace{-0.4cm}
  \caption{(a) AR with different numbers of hypotheses. (b) AR of final pose and HR of pose hypotheses with different numbers of correspondence samples. }
  \label{fig:ablation}
  \vspace{-0.6cm}
\end{figure}

\noindent\textbf{Influence of the Number of Hypotheses.}
As shown in Figure \ref{fig:ablation}~(a), our method benefits from increasing the number of hypotheses. However, the performance saturates when the number passes 50, indicating 50 hypotheses could cover true poses for most cases. To balance the computation cost and performance, we use 20 as the default.

\noindent\textbf{Quality of Pose Hypotheses.}
To investigate the quality of our pose hypotheses, we define the Hit Recall (HR) as the ratio of testing set whose ground truth pose is included in our proposed hypotheses. Specifically, the top 128 correspondences are used as the sampling pool, and one hypothesis is computed through 3 correspondences randomly selected from it. This procedure is repeated to generate multiple hypotheses. We report the HR in comparison with the Average Recall ~(AR) by varying the number of correspondence samples. As shown in Figure~\ref{fig:ablation}~(b), the AR exhibits a lower number since it only considers the top-1 scored pose hypothesis. When considering all the hypotheses in evaluation, our method achieves over 80\% HR, which reflects the potential of our descriptors in generating the correct poses.


\section{Conclusion}
\label{sec:conclusion}
We present \OURS{}, a Fuse-Describe-Match framework for unseen object pose estimation from single RGB-D images.
Our method first extracts rotation-invariant descriptors from 3D point clouds of CAD model and depth map. Then, the multi-modal fusion of texture and geometry is achieved through a Latent Fusion Attention Module. A Bridged Coarse-Level Matching Loss is introduced to utilize latent features from RGB images to connect descriptions of partial observations and full object geometry. \OURS~inherently models object symmetry without explicit annotations. 
MatchU surpasses all existing methods for unseen object pose estimation by a large margin on standard benchmarks. Certainly, it relies on external object localization and could be could be affected by their erroneous results. In the future, incorporating such modules into the pipeline to build end-to-end training might further improve our results. We believe that by closing the gap to object-specific baselines, \OURS~constitutes an important step forward to truly scalable 6D pose estimation of unseen objects.


{
\small
\bibliographystyle{ieeenat_fullname}
\bibliography{main}
}

\end{document}